\documentclass[letterpaper, 10 pt, conference]{ieeeconf}  

\IEEEoverridecommandlockouts                              

\title{\LARGE \bf
End-to-end Learning of Cost-Volume Aggregation for\\ Real-time Dense Stereo.
}

\usepackage{graphicx}
\usepackage[]{algorithm2e}
\usepackage{amsmath}
\usepackage{mathtools}

\newcommand{\fig}[1]{Figure~\ref{fig:#1}}

\newcommand{\eq}[1]{(\ref{eq:#1})}

\newcommand{\ignore}[1]{}

\begin{document}



\author{Andrey Kuzmin \\ Skolkovo Institute of Science and Technology \\ Dmitry Mikushin \\ Applied Parallel Computing LLS \\ Victor Lempitsky\\ Skolkovo Institute of Science and Technology}

\maketitle

\begin{abstract}
We present a new deep learning-based approach for dense stereo matching. Compared to previous works, our approach does not use deep learning of pixel appearance descriptors, employing very fast classical matching scores instead. At the same time, our approach uses a deep convolutional network to predict the local parameters of cost volume aggregation process, which in this paper we implement using differentiable domain transform. By treating such transform as a recurrent neural network, we are able to train our whole system that includes cost volume computation, cost-volume aggregation (smoothing), and winner-takes-all disparity selection end-to-end. The resulting method is highly efficient at test time, while achieving good matching accuracy. On the KITTI 2015 benchmark, it achieves a result of 6.34\% error rate while running at 29 frames per second rate on a modern GPU.
\end{abstract}

\section{Introduction} \label{section_intro}

The use of deep convolutional networks has recently advanced the accuracy of stereo matching algorithms considerably \cite{zbontar2015stereo,luo2016efficient,zbontar2016stereo,mayer16}. This improvement has been facilitated by the emergence of sizeable training sets, such as the KITTI datasets for autonomous driving~\cite{Geiger2012CVPR}, the new version of the Middlebury dataset~\cite{scharstein2014high}, and, most recently, synthetic datasets of high quality \cite{RosCVPR16}. The use of machine learning allows to tune the stereo matching process to handle characteristic image patterns. This allows to resolve various stereo ambiguities using semantic cues, surpassing the accuracy of more traditional approaches that use low-level cues and priors.  

Despite this success of deep learning methods in stereo, designing real-time algorithms as required by the majority of applications has proved challenging. The initial approach of \cite{zbontar2015stereo} required over a minute to process a KITTI stereo-pair. A more recent (``fast'') variant discussed in the follow-up work~\cite{zbontar2016stereo} and a similar approach of \cite{luo2016efficient} have brought down this time to as little as one second, which is still excessive for many applications. 

The computational bottleneck within methods \cite{zbontar2015stereo,zbontar2016stereo,luo2016efficient} is in the matching process of high-dimensional descriptors of local appearance, which has to be done for all pairs of potentially matching pixels across the two views. The most recent method \cite{mayer16} streamlines this necessity by proposing a deep architecture that directly outputs the disparity map given the stereo-pair as an input. While the high-dimensional descriptors still have to be implicitly matched within their architecture, this matching process only happens at low resolution, while further processing results in the efficient upsampling.

Here, we propose a new way to apply deep learning in order to improve the accuracy of stereo matching. In order to achieve real-time frame rate, we avoid learning high-dimensional descriptors and matching them and focus our learning-based effort on the cost-aggregation process. We thus use a simple linear combination of two classical and very fast similarity measures based on census transform \cite{zabih1994non} and sum-of-absolute-differences matching to define the overall matching costs for various pixels and disparities. 

To perform cost-aggregation, we smooth the obtained noisy matching costs using one of the fastest edge-preserving smoothing techniques, namely \textit{domain transform}~\cite{gastal2011domain,pham2013domain} across the four directions. Crucially, we make the parameters of this cost-aggregation process spatially-varying and use a deep convolutional network to predict them on a per-pixel basis. Such prediction facilitates smoothing across parts belonging to same object and prevents smoothing across object boundaries. At test time, the deep learning module processes only one of the input images, and the complexity of this module is thus independent of the disparity range.

Our experiments demonstrate that a combination of a simple matching process and a trainable domain transform-based cost aggregation is able to achieve a unique combination of a high frame-rate (e.g.\ 29 fps for KITTI 2015 dataset) and high matching accuracy (state-of-the-art for real-time methods). The high accuracy is obtained via the end-to-end learning process that takes into account the pixel-level matching, the cost aggregation, and the final winner-takes-all disparity selection. The ultimate accuracy greatly benefits from initializing the weights within our deep network to the weights of the network trained to detect natural boundaries in images~\cite{xie2015holistically}. 

In the remainder of the paper, we discusses the related work (\ref{section_related}), detail the proposed method (\ref{section_method}), present the results of the experimental validation in \ref{section_results}, and conclude with a short discussion and outlook (\ref{section_conclusion}). 

\section{Related work} \label{section_related}

Our work is related to a large body of works on fast stereo matching that investigate the use of efficient algorithms such as few rounds of message passing~\cite{hirschmuller2005accurate}, bilateral filtering~\cite{yoon2006adaptive}, guided filtering~\cite{rhemann2011fast}, or domain transform~\cite{pham2013domain} in order to achieve smoothness in the reconstructed depth maps. The domain transform used in our approach and introduced in \cite{gastal2011domain} can be regarded as a fast approximation to bilateral filtering and is overall the fastest of the global aggregation methods employed within this class of stereo methods.

Our approach has been inspired by the recent work of \cite{chen2015semantic} that establishes the connection between the domain transform and the gated recurrent neural networks (such as LSTM \cite{hochreiter1997long} and GRU \cite{cho2014learning}) and then uses this connection to discriminatively train domain transform parameters for the task of semantic segmentation. Our approach is reminiscent of other works on semantic segmentation such as \cite{zheng2015conditional} and \cite{schwing15} that also draw connections between recurrent neural networks and conditional random fields, and use these connections to impose spatial smoothness. Very recently (and concurrently) to our work, several groups have considered the use of learnable edge-aware smoothing for several image processing operations, including post-processing of depth maps \cite{Barron16,Liu16}.

As discussed above, our work is also related to preceding approaches that use deep learning for stereo. Our approach differs markedly from \cite{zbontar2015stereo,zbontar2016stereo,luo2016efficient} as we use deep learning within the cost aggregation rather than to compute the matching costs themselves. Unlike~\cite{zbontar2015stereo,zbontar2016stereo,luo2016efficient} and similarly to \cite{mayer16} we also use end-to-end learning that encompasses all stages of depthmap computation within our method. Unlike \cite{mayer16}, which uses a rather generic feed-forward convolutional network trained on a massive amount of synthetic stereo pairs, our method employs classical stereo matching algorithms such as census transform as modules within a more specific architecture that combines convolutional networks with a gated recurrent neural network module (which is equivalent to the domain transform operation).

\section{Method} \label{section_method}

We consider a dense stereo correspondence problem where a rectified pair of images $I^L$ and $I^R$ (the left and the right view images respectively) is given as an input and the goal is to assign each pixel $(x,y)$ in the left image $I^L$ a label $d$ from the set $d \in [0,d_{max}]$ where $d_{max}$ is the maximum allowed disparity. Our approach consists of three steps: constructing the cost volume, the cost-volume aggregation and the winner-takes-all label selection. As a postprocessing we also apply left-right consistency check to identify and fill in the occluded parts. We now discuss these steps in detail.

\subsection{Computing stereo-matching costs}

In our method we operate on the cost volume explicitly stored as a three-dimensional array with dimensions $(h,w,d_{max})$ where $h$ and $w$ are the image dimensions.

Following the pipeline of a typical local stereo method, we compute the following stereo matching cost based on two terms. The first term is based on the sum of absolute differences (SAD). We use 1x1 patches (i.e.\ individual pixels) for the sake of preserving maximum amount of texture information and rely on the subsequent cost aggregation scheme for further smoothing:
\begin{equation}
E_{SAD}(x,y,d) = \sum_{r,g,b}|I^{L}(x,y)-I^{R}(x-d,y)|
\end{equation}
The second term $E_{census}(x,y,d)$ is based on matching of the local census transform \cite{zabih1994non} which is a non-parametric local transform that relies on the relative order of intensity values. We convert images to gray-scale to compute this term. The local image structure at each $n\times n$ patch is summarized by $(n^2-1)$ binary bits. Each bit is set according to the comparison of the patch's central pixel intensity with the remaining pixels of the patch. The obtained descriptors are then matched with the hamming distance when computing the stereo matching costs. The census transform matching computation can be naturally mapped to a data parallel pipeline thus making it very efficient for modern GPU architectures. Each thread is assigned to a single pixel computation which is performed in constant time.
Finally we combine the two terms into the final matching cost:
\begin{equation}
E(x,y,d) = \alpha E_{SAD} (x,y,d) + (1-\alpha) E_{census} (x,y,d)
\end{equation}
Where $\alpha \in (0,1)$ and is a constant coefficient that controls the ratio between the two cost values.  

\subsection{Spatial boundary detector}

\begin{center}
\begin{figure*}
\centering
\begin{small}
\begin{tabular}{c}
\includegraphics[width=17.0cm]{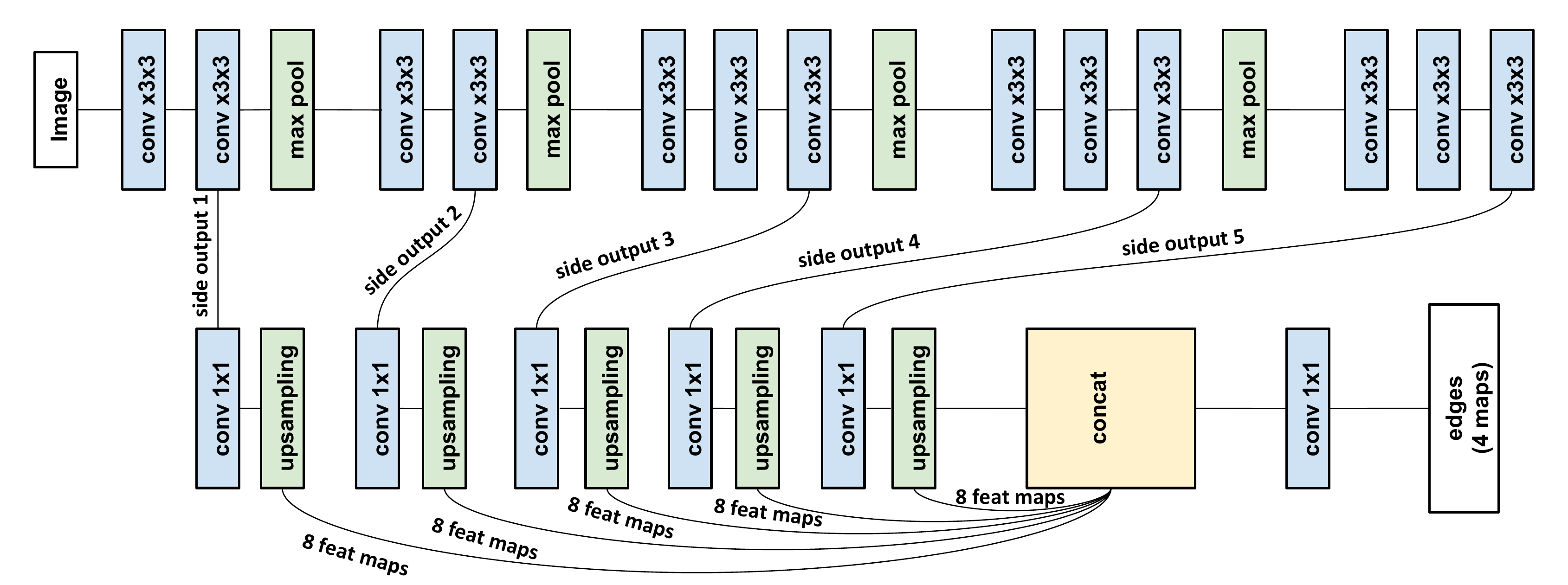}
\end{tabular}
\end{small}
\caption{Architecture of the HED edge detector \cite{xie2015holistically}. The last convolutional layers at each scale are connected directly to the final edge map using the side outputs. Thus the supervision guides the side outputs towards edge predictions at the corresponding scale. The amount of feature maps has been increased from 1 to 8 at each of the upsampling layers in order to produce 2 diverse feature maps at the last convolutional layer, each is used to the horizontal and the vertical pass of the domain transform.}
 
\label{fig:edge_detector_arch}
\end{figure*}
\end{center}
Similar to \cite{rhemann2011fast} we perform cost-volume aggregation using a spatially-varying smoothing process. We rely on deep convolutional network and the end-to-end training to set the smoothing in an optimal and problem-specific way. Thus, our task is to go beyond simple edge detection, as not all the edges on the images are aligned with the disparity transitions. Therefore, the accuracy of the method could increase during the supervised training. 

Still, because of an inherrent connection between object boundaries and edge discontinuities, our method for smoothing weight computations is based on the CNN-based architecture for object boundary detection proposed in \cite{xie2015holistically}. This architecture is thus embedded into the end-to-end learning process. The method \cite{xie2015holistically} addresses the challenging ambiguity in edge and object boundary detection by learning rich hierarchical representation. It combines the edges detected at different image scales into a final edges map. To achieve this combination, their architecture includes skip connections merging together edges predicted at multiple scales (see fig. \ref{fig:edge_detector_arch}). Thus the final loss is informed about the edges predictions from the range of five scales.

Since our stereo method requires prediction of two edge maps (see the following sections for details), we slightly modify the original architecture in the following way. The amount of feature maps at each of the side output is increased from one to eight, each of them is further upsampled, so that the last 1x1 convolutional layer is able to produce two-channel output.

\subsection{Smoothing with a domain transform}

\begin{figure*}
\centering
\begin{small}
\begin{tabular}{cccc}
\includegraphics[width=3.9cm]{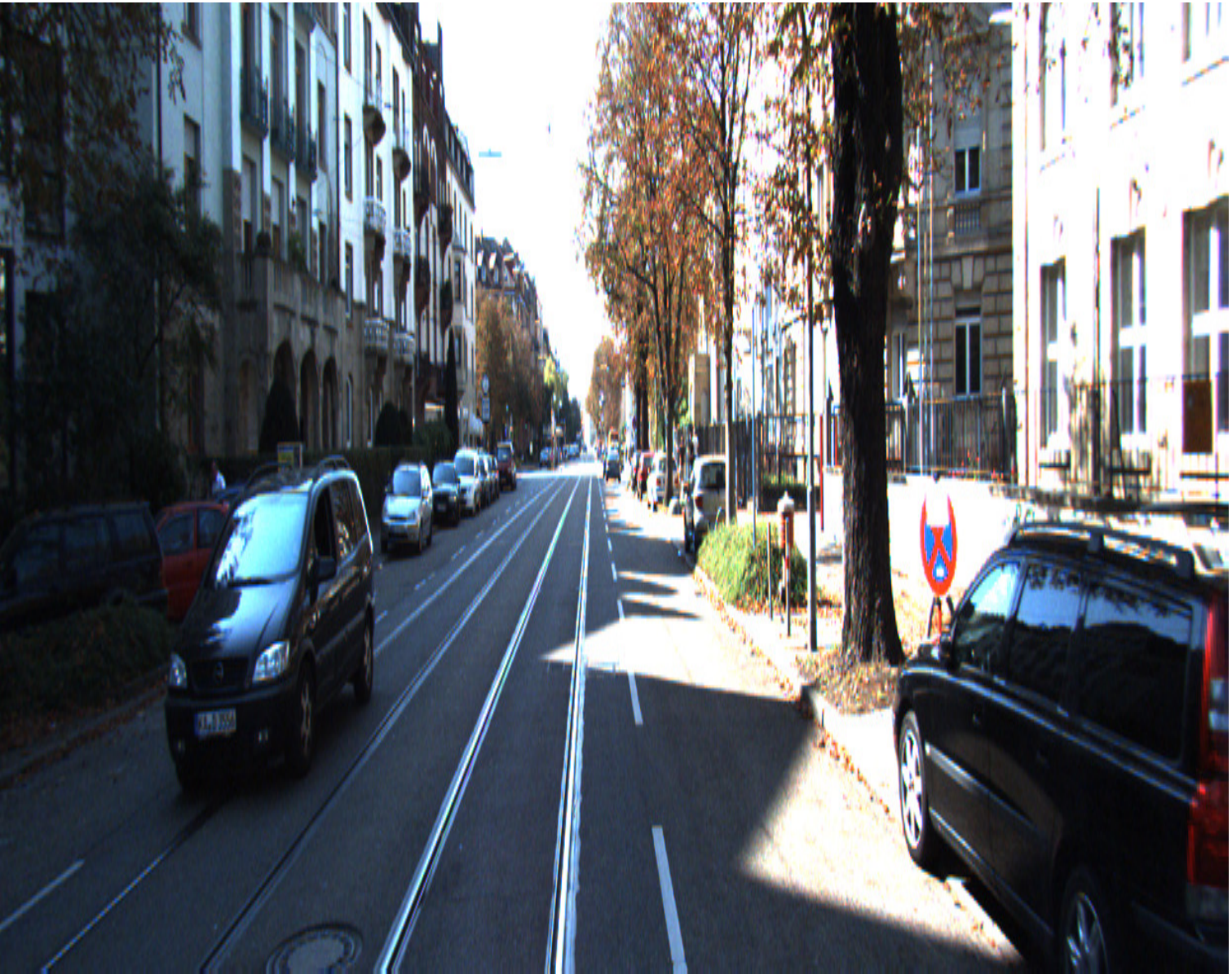}&
\includegraphics[width=3.9cm]{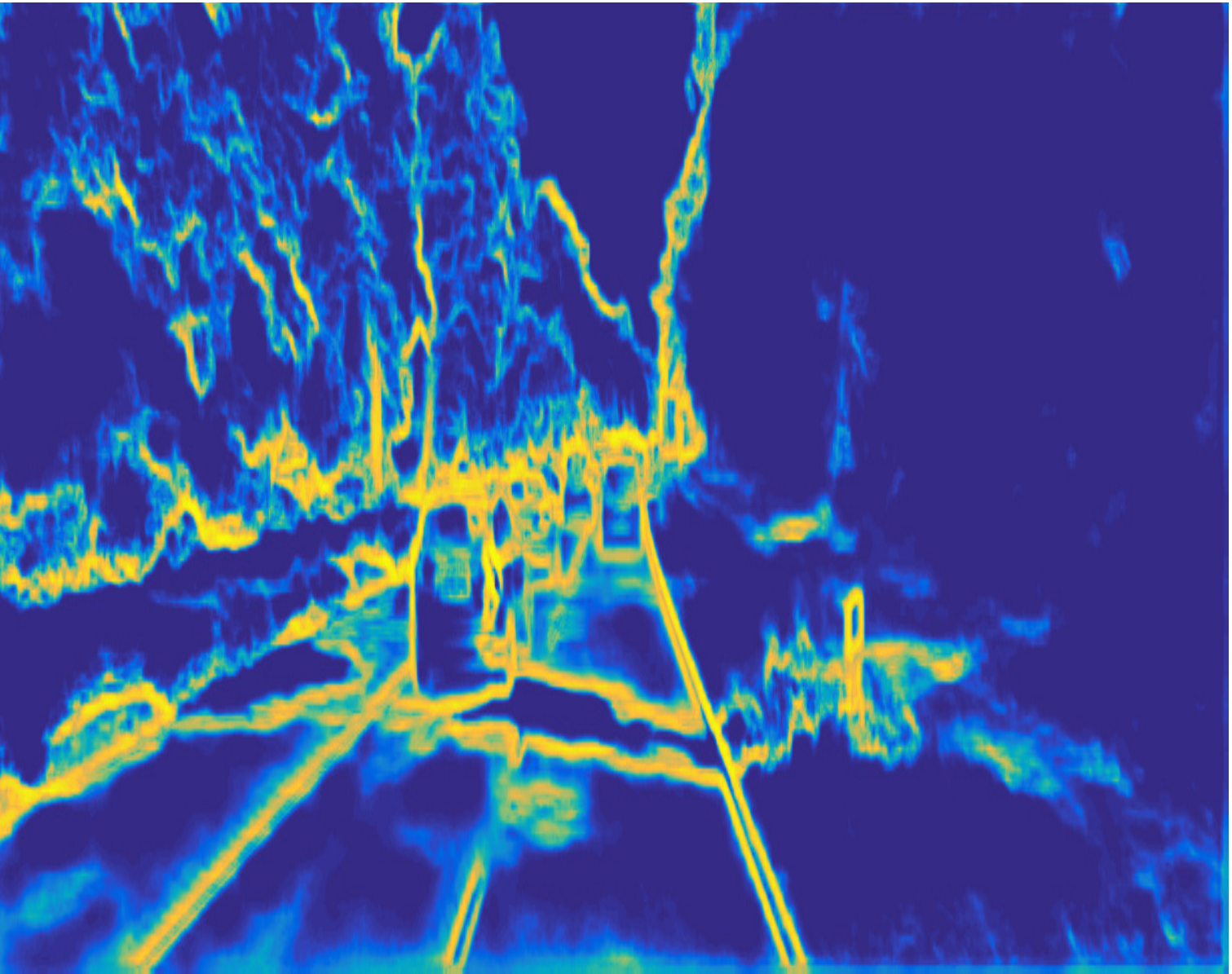}&
\includegraphics[width=3.9cm]{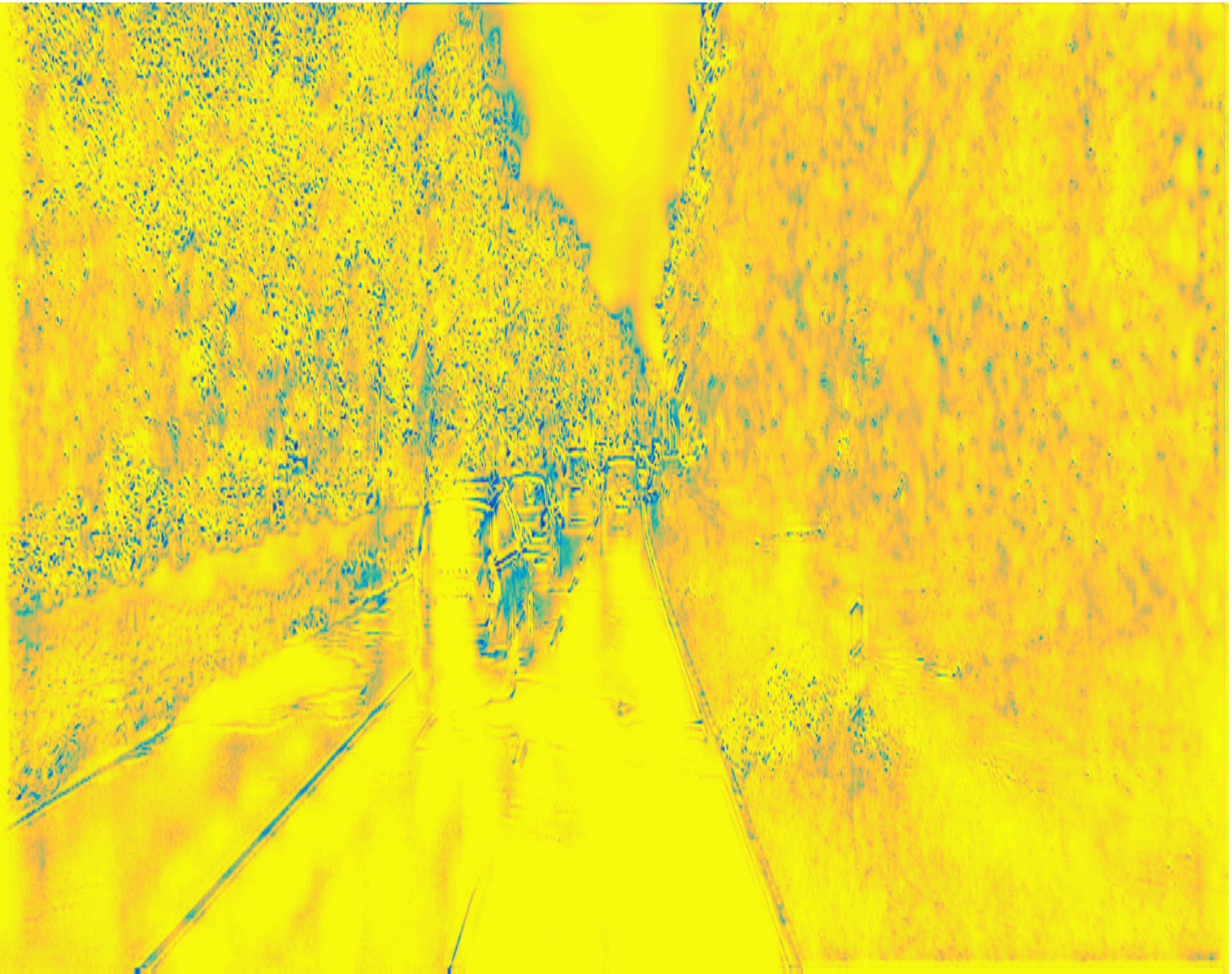}&
\includegraphics[width=3.9cm]{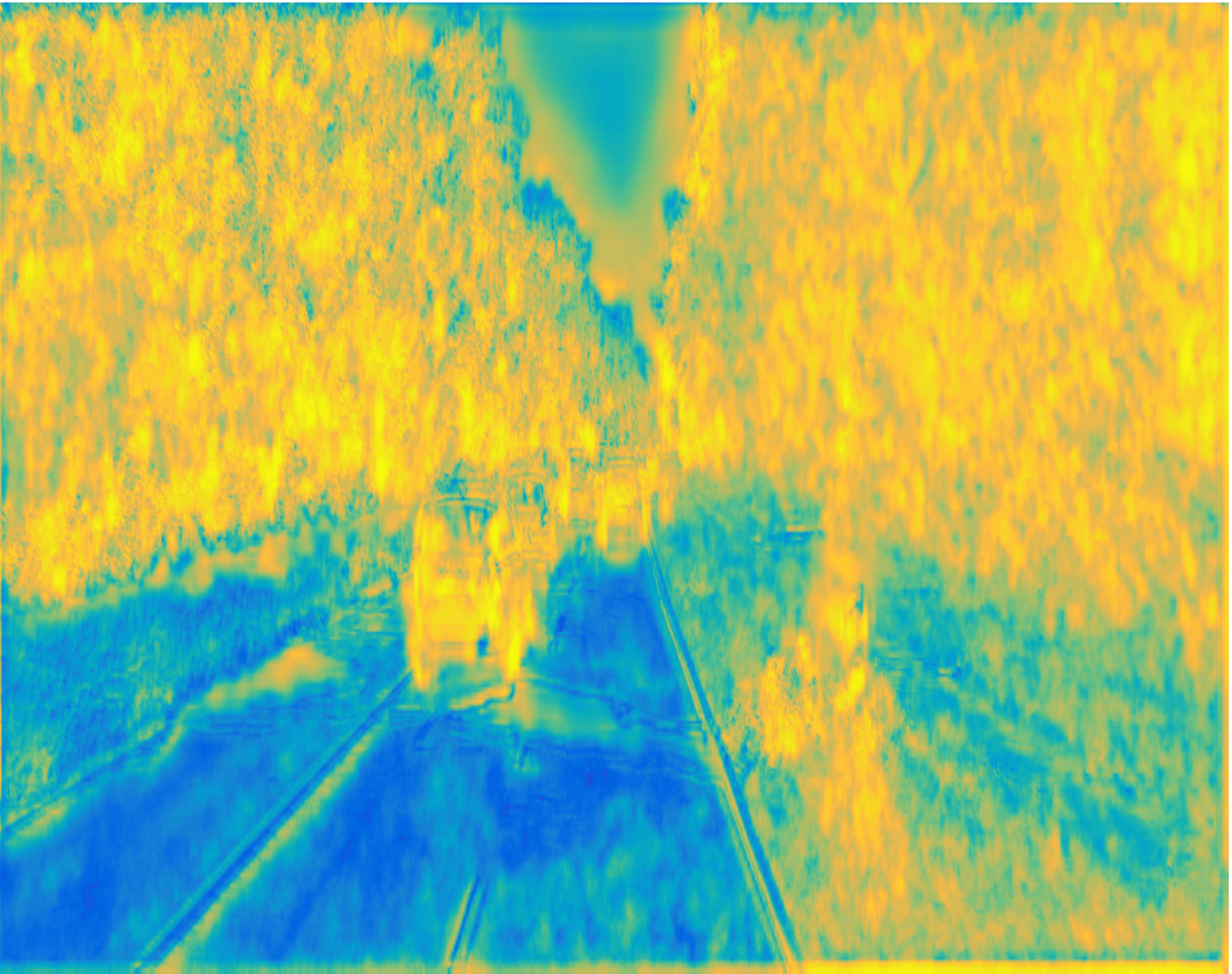}
\includegraphics[trim=400 0 0 0,clip,height=3.15cm]{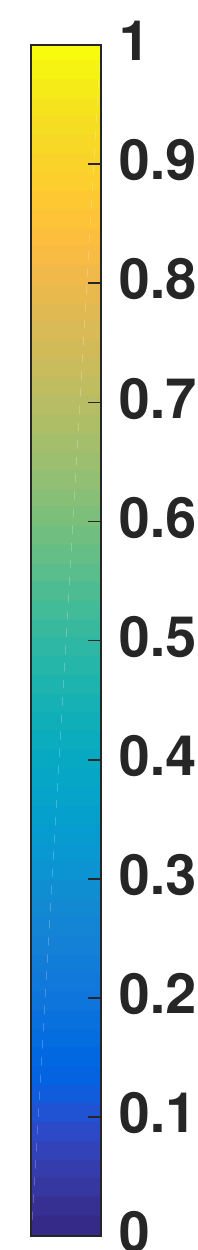}
\end{tabular}
\end{small}
\caption{Example of learned domain transform weights. Original image (1), initial weights map (2), horizontal pass weights (3), vertical pass weights (4). The set of edges detected by the HED model differs significantly compared to the original output.}
\label{fig:learned_weights}
\end{figure*}

Our cost-volume aggregation scheme is based on the domain transform method originally proposed for the edge-preserving image filtering \cite{gastal2011domain} and further used in a neural network pipeline \cite{chen2015semantic} in the context of semantic segmentation.

The domain transform method is originally used in order to perform fast edge-aware filtering of an image guided by a gradient information. It can be regarded as an instance of a fast bilateral filter \cite{gastal2011domain}. The domain transform takes two inputs:(1) the input signal $x$, and (2) is the map of weights $w_{i}\in[0,1]$. The output of the domain transform is a filtered signal $y$. For a 1-D signal the output is given recursively by the following recursive relation. After setting $y_1=x_1$, for $i=2,...,N$.

\begin{equation}
\label{eq:dt_forward}
y_i = (1-\omega_{i})x_{i} + \omega_{i}y_{i-1}    
\end{equation}

The set of weights $\omega_i$ is used in order to control the amount of smoothing along the signal yielding a way to preserve the edges by controlling the magnitude of $w_i$. Indeed in the regions of $\omega_i$ close to one, the maximum amount of information propagates from the previous pixel $y_{i-1}$ to the current pixel $y_i$. On the contrary, if $w_i$ is small (e.g.\ in the regions of large signal gradient), the output equals to the input, i.e.\ $y_i=x_i$. 

Domain transform filtering for 2-D images works in a separable way, using 1-D filtering sequentially along each dimension: a horizontal pass (left-to-right and right-to-left) along each row is followed by the vertical pass along each column (top-to-bottom and bottom-to-top). For the reasons described in the subsection \ref{subsec_costvol_filt} we use distinct weight maps $W_{hor}$ and $W_{vert}$ for the horizontal and the vertical passes respectively. We introduce the following notation for a 2D domain transform which takes an image $I$ and the two weight maps and computes the filtered image $I_{filt}$:

\begin{equation}
I^{filt}=\text{DT}(I,W_{hor},W_{vert})
\end{equation}

The procedure can be formally described as the sequence of the four recurisive passes (left-right, right-left,  top-bottom, bottom-top) computed in the following order:

\begin{equation}
\begin{split}
I^L(x,y) = (1-W_{hor}(x,y))\,I(x,y) + \\ W_{hor}(x,y)\,I^L(x-1,y) 
\end{split}
\end{equation}

\begin{equation}
\begin{split}
I^R(x,y,d) = (1-W_{hor}(x,y))\,I(x,y) + \\ W_{hor}(x,y)\,I^R(x+1,y) 
\end{split}
\end{equation}

\begin{equation}
\begin{split}
I^T(x,y,d) = (1-W_{vert}(x,y))\,I(x,y) + \\
W_{vert}(x,y)\,I^T(x,y-1) 
\end{split}
\end{equation}

\begin{equation}
\begin{split}
I^B(x,y,d) = (1-W_{vert}(x,y))\,I(x,y) + \\ W_{vert}(x,y)\,I^B(x,y+1) 
\end{split}
\end{equation}

In order to achieve our goal, the weights must be predicted based on the input image. The authors of \cite{chen2015semantic} proposed a piecewise-differentiable version of the domain transform in order to backpropagate the errors within the semantic segmentation pipeline. This approach is extended to the task of cost volume filtering as described in the following section. 

In order to explain how backpropagation works for the 1D filtering process of \eq{dt_forward} we assume the output is given as an input to the subsequent layer $L$. So each sample $y_i$ of the output signal receives contribution of the gradients $\frac{\partial L}{\partial y_i}$. In order to compute gradients of the inputs, we unroll the recurrence \eq{dt_forward} in the reverse order, i.e.\ for $i=N,N-1,...,2$:

\begin{equation}
\frac{\partial L}{\partial x_i} = (1-\omega_i)\frac{\partial L}{\partial y_i}
\end{equation}

\begin{equation}
\frac{\partial L}{\partial w_i} = \frac{\partial L}{\partial w_i} + (y_{i-1}-x_i)\frac{\partial L}{\partial y_i}
\end{equation}

\begin{equation}
\frac{\partial L}{\partial y_{i-1}} = \frac{\partial L}{\partial y_{i-1}} + \omega_{i}\frac{\partial L}{\partial y_i}
\end{equation}

Thus, the four passes of the domain transform can be combined into a learning pipeline where the recursive relation (\ref{eq:dt_forward}) can be considered as an instance of the gated recurrent unit~\cite{chen2015semantic}. \ignore{TODO: give some details and reference the GRU paper} Equation (\ref{eq:dt_forward}) defines DT filtering as a recursive operation. In fact, in fact there is a precise connection to GRU which was recently proposed for modelling sequential data \cite{cho2014properties}. The value $(1-w_i)$ is related to GRU's "update gate" and $y_{i-1}$ is a "candidate activation"~\cite{chen2015semantic}. 

\subsection{Cost-volume filtering}
\label{subsec_costvol_filt}
\begin{center}
\begin{figure*}
\centering
\begin{small}
\begin{tabular}{c}
\includegraphics[width=11.0cm]{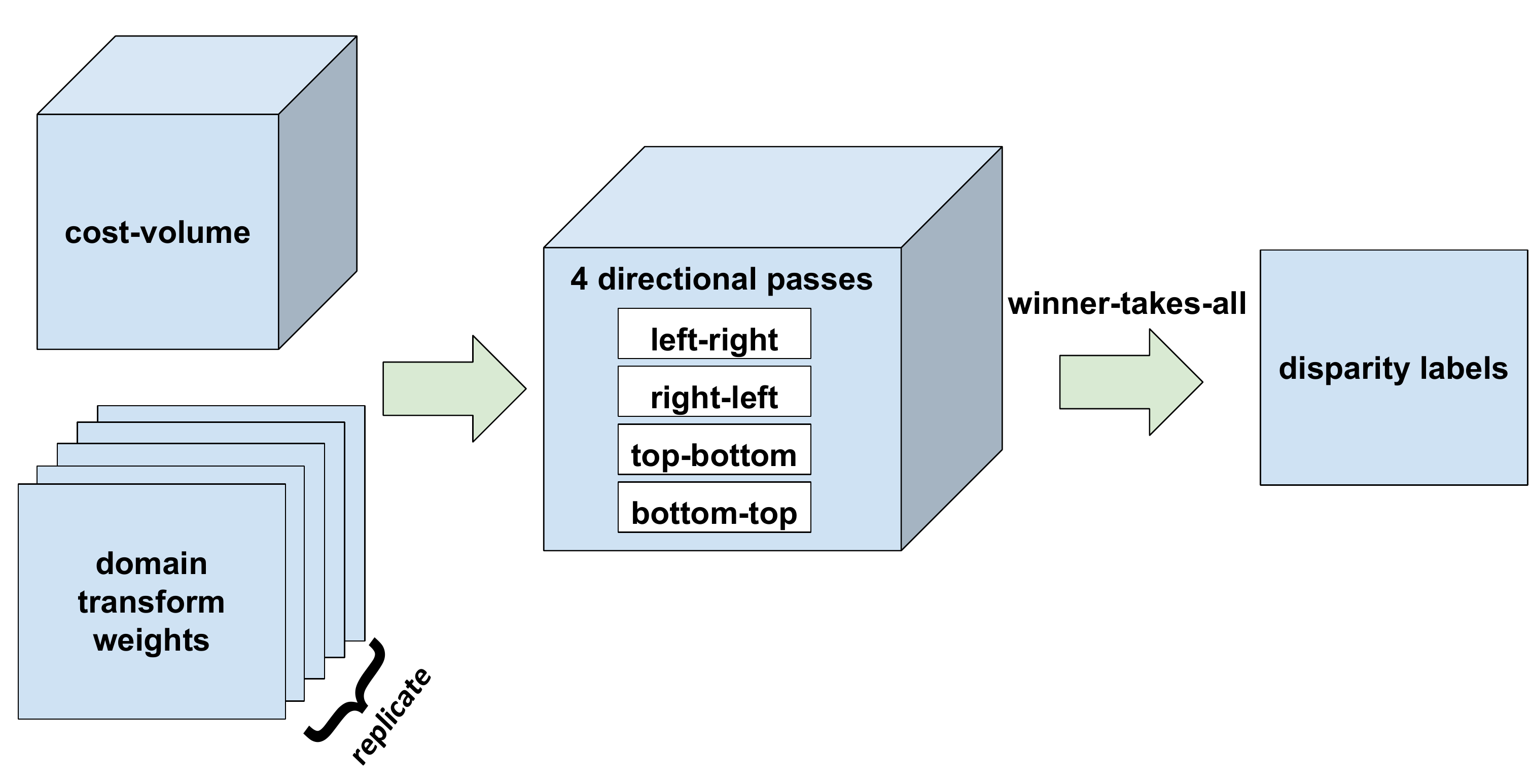}
\end{tabular}
\end{small}
\caption{Cost-volume filtering algorithm schematic view. The domain transform weights are replicated for each slice of the matching cost volume and 4 directional passes (left-right, right-left, top-bottom, bottom-top) are performed in parallel. Finally, winner-takes-all strategy is used for the disparity labels selection.}
 
\label{fig:cost_aggr}
\end{figure*}
\end{center}

We follow the idea of cost-volume smoothing \cite{rhemann2011fast} using an edge-preserving filter. In fact our approach might be considered as the generalization of \cite{pham2013domain,cigla2015recursive} to a machine learning pipeline where the filter weights are predicted by the convolutional neural network.

The overall scheme of our algorithm is given in \fig{edge_detector_arch}. 
The cost volume filtering is performed in four directional passes of the domain transform. In order to use two-dimensional domain transform weights $W_{hor,vert}(x,y)$ for the three-dimensional array filtering, we simply replicate the 2D edge maps for each slice of the energy tensor. For each slice of the cost volume $E_{d} = E(x,y,d)$, $x=0,..,w$, $y = 0,..,h$, the domain transform is computed independently as follows: \ignore{TODO(done): ADD SOME FORMULAS HERE:}

\begin{equation}
\begin{split}
E^{filt}_d=\text{DT}(E_d,W_{hor},W_{vert}), \\ \text{$d = \{0,1,...,{d_{max}}\}$}  
\end{split}
\end{equation}

The cost-volume filtering is than followed by the classical winner-take-all strategy for disparity label selection.

The authors \cite{chen2015semantic} use the shared weights $w_{ij}$ for each of the four directional passes. We, however, use separate weight maps for the vertical and the horizontal passes $W_{hor,vert}$ as it is beneficial for our task due to the the fact that the rate and the statistics of disparity changes along the horizontal and the vertical directions can be quite different. Using two weight maps instead of one implies very small computational overhead during the test-time stereo processing, while increasing the accuracy.

In order to reduce the computational complexity of our algorithm we train the edge detector for the half resolution of the original image and then use bilinear interpolation in order to upsample the edge maps to the original size. While the cost volume is computed for the original image size. Indeed, there is sufficient amount of information in real-world imagery at half resolution in order to extract edges relevant to the disparity discontinuities. Thus, some additional run-time reduction is gained at the cost of little or no increase in the disparity error.

We use the exponential non-linearity to map the output of the convolutional network to the weights of the domain transform:
\begin{equation}
W_{vert,hor} = \exp (-\sigma E_{vert,hor})
\end{equation}
where $\sigma$ is, once again, a tunable parameter, that affects the convergence speed of the training process.

\subsection{Loss}

In order to match the filtered cost volume with the ground truth disparity field, we represent each of the ground truth labels as a one-hot vector and use the cross-entropy soft-max loss to evaluate the final disparity field error, thus maximizing the log-probability of the correct displacement at every pixel (where the disparity is known at training time). We did not observe practical benefit from using narrow gaussian distribution instead of one-hot vector as proposed in \cite{luo2016efficient}.

\begin{center}
\begin{figure*}
\centering
\begin{small}
\begin{tabular}{c}
\includegraphics[width=17.0cm]{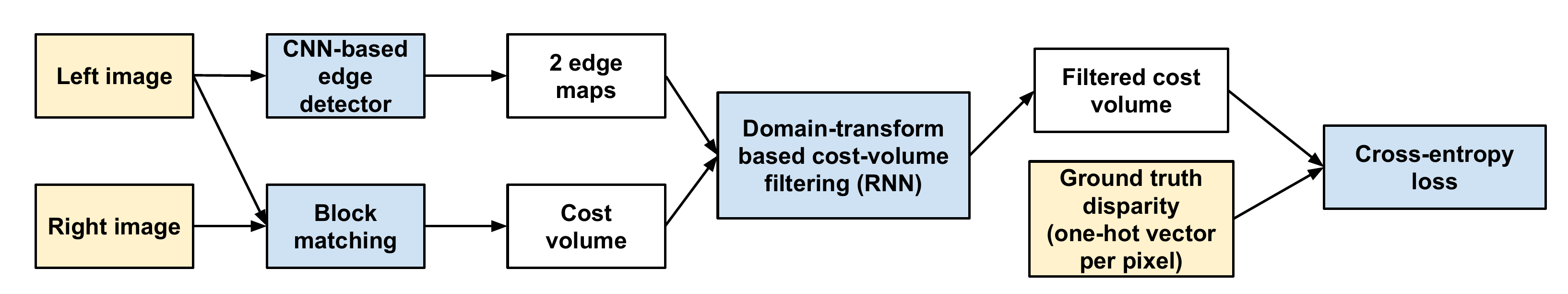}
\end{tabular}
\end{small}
\caption{The overall scheme of our learning pipeline. The left and the right images are given as an input for the raw matching cost volume. The left image chosen as the reference view for the disparity computation serves as an input to the edge detection CNN. The domain transform which can be regarded as an instance of RNN takes two inputs: the raw cost volume and the pair of edge maps predicted by the CNN. Finally, the cross-entropy loss is used to penalize the error between the predicted disparity and the ground-truth values. The error is back-propagated to the CNN values through the domain-transform RNN.}
 
\label{fig:pipeline}
\end{figure*}
\end{center}

The training in our method is end-to-end (fig.~\ref{fig:pipeline}) as the disparity field error is backpropagated through the recurrent part (comprised of four differentiable domain transforms) to the weight-computing convolutional neural network, so that the network learns to find disparity edges suitable for the cost volume filtering.

\section{Experiments} \label{section_results}


\begin{center}
\begin{figure*}
\centering
\begin{small}
\begin{tabular}{c}
\includegraphics[width=10.0cm]{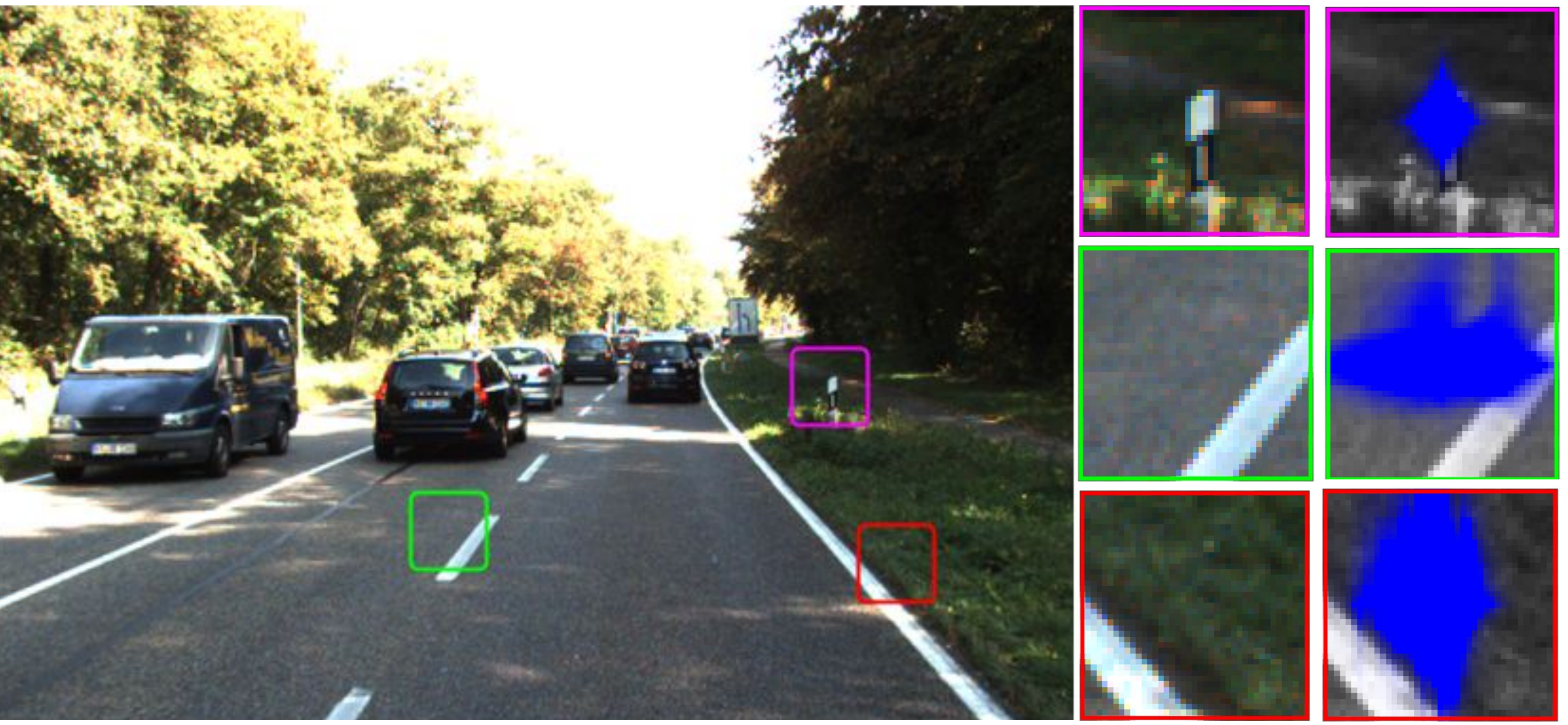}
\end{tabular}
\end{small}
\caption{Example of the filtering process with the trained domain transform. Each of the three patches on the right hand side denoted with magenta, green and red colors respectively show relative contribution of its central pixel to the rest of the pixels of the patch. The central pixel of the first patch shown in magenta corresponds to the road milestone, thus the smoothing process is localized by the border of the milestone. The patches shown in green and red demonstrate how the road markings relevant (red) and irrelevant (green) to the disparity change are discriminated by the trained domain transform.}
 
\label{fig:dt_viz}
\end{figure*}
\end{center}

\subsection{Data set}

We evaluated our method on the KITTI 2015 public data set \cite{menze2015cvpr}., which is a collection of color image pairs taken from a car roof while driving in a European city. Each of the pair rectified and the ground truth is given in a form of a sparse disparity field obtained with LIDAR. The disparity field regions corresponding to cars were then refined to dense fields using geometrically accurate CAD models fit to the point clouds.
Each image has dimensions $1242 \times 375$ pixels with relatively large displacement within $[0..256]$ pixels range. Computing the stereo correspondence is especially challenging around the reflecting surfaces especially car windshields and windows, textureless regions including homogeneous car bodies, and traffic forming thin regions surrounded by large disparity fields discontinuities. Whereas most of the disparity fields consist of slanted surfaces corresponding to the road surface.     

\subsection{Left-right disparity check.}

Since the ground truth disparity fields include occluded regions, it is necessary to perform the standard left-right check procedure in order to interpolate the disparity fields within those regions in order to obtain accurate results. Following the algorithm described in \cite{zbontar2015stereo}, we compute the disparity fields for both left-right and right-left image pairs and then interpolate the occluded and mismatched pixels.

\subsection{Details or learning}
The data set was split into the training set (160 image pairs) and the validation set (40 images). 
The pre-processing including mean subtraction was adopted from the edge detection network. 

The data term corresponds to the linear combination of $1 \times 1$ SAD difference blocks with $7 \times 7$ census transform, the value parameter $\alpha=0.43$ is chosen experimentally. The parameter $\sigma$ is set to $4$ and does not affect the final accuracy.

The original architecture was modified in two ways. First, the amount of feature maps in the convolutional layers was halved in order to decrease the run-time. Second, the amount of feature maps used for the final linear combination was changed from 1 to 8 (fig. \ref{fig:edge_detector_arch}). 

We train the network to minimize the cross-entropy loss using ADAM \cite{kingma2014adam} method. The learning rate is set to $2.5\cdot10^{-5}$.
We benefit from using the edge detection model pretrained to extract edges for BSDS data set \cite{MartinFTM01}. The network is pre-trained for a edge-detection task. The ground truth labels obtained from the manual annotation are provided. The combination of cross-entropy losses for different scales is performed as described in \cite{xie2015holistically}. \ignore{TODO(DONE): DEVOTE A SEPARATE PARAGRAPH TO PRE-TRAINING, MAYBE HERE, MAYBE IN THE END OF THE METHODS SECTION.}

An example of the original HED edge detection output compared to the learned domain transform weights can be observed on the fig. \ref{fig:learned_weights}. Although there is no general intuition on the learned weights, one can observe that some of the edges were suppressed during the training. The degree of smoothing for the horizontal and the vertical pass is different after the training.

\begin{figure*}
\centering
\begin{small}
\begin{tabular}{ccc}
\includegraphics[width=5.0cm]{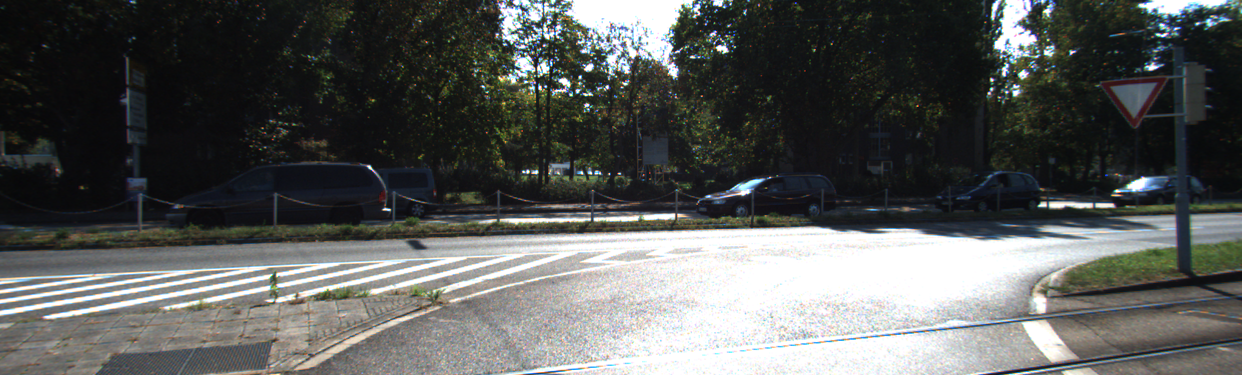}&
\includegraphics[width=5.0cm]{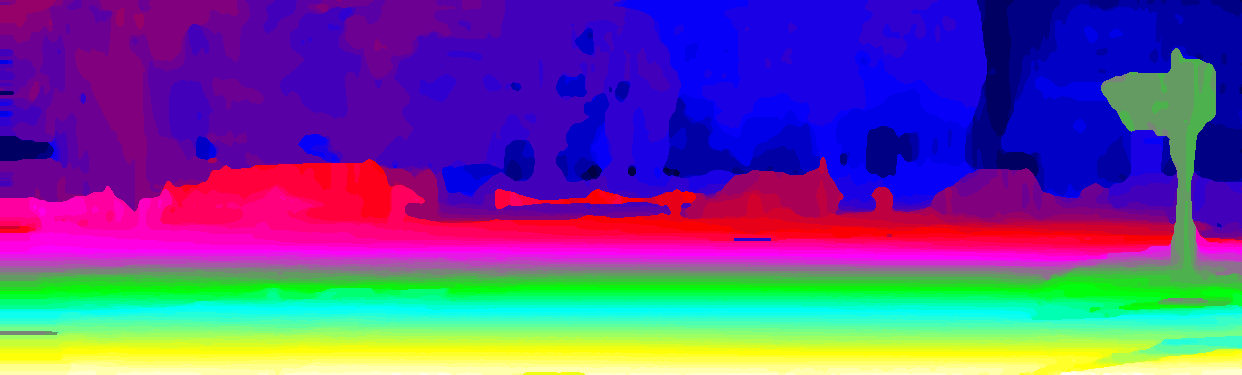}&
\includegraphics[width=5.0cm]{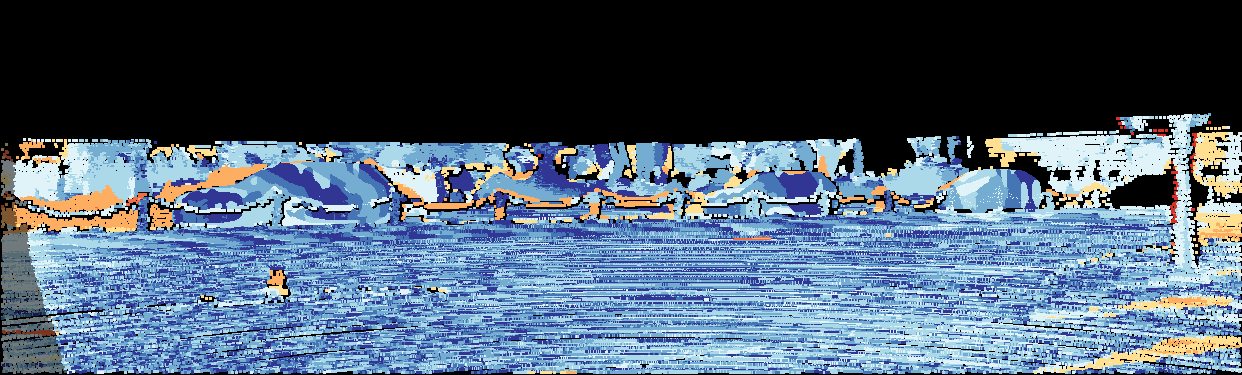}\\

\includegraphics[width=5.0cm]{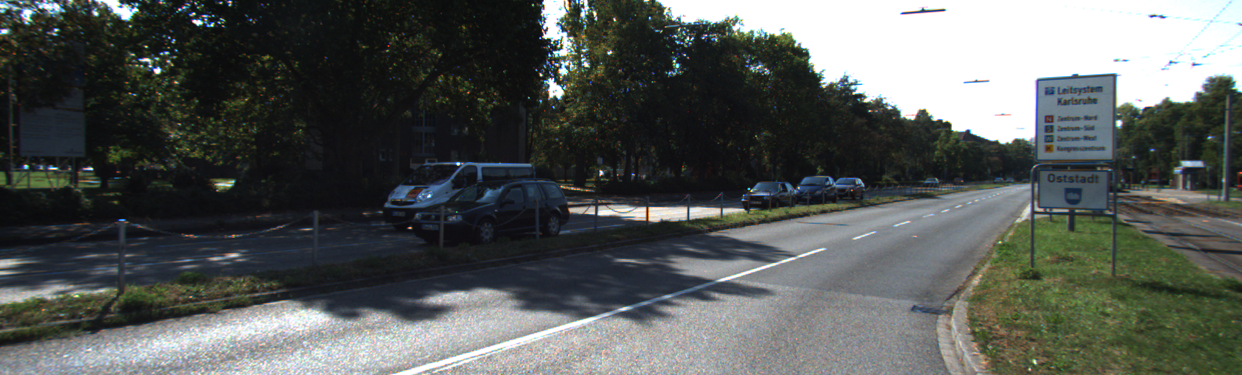}&
\includegraphics[width=5.0cm]{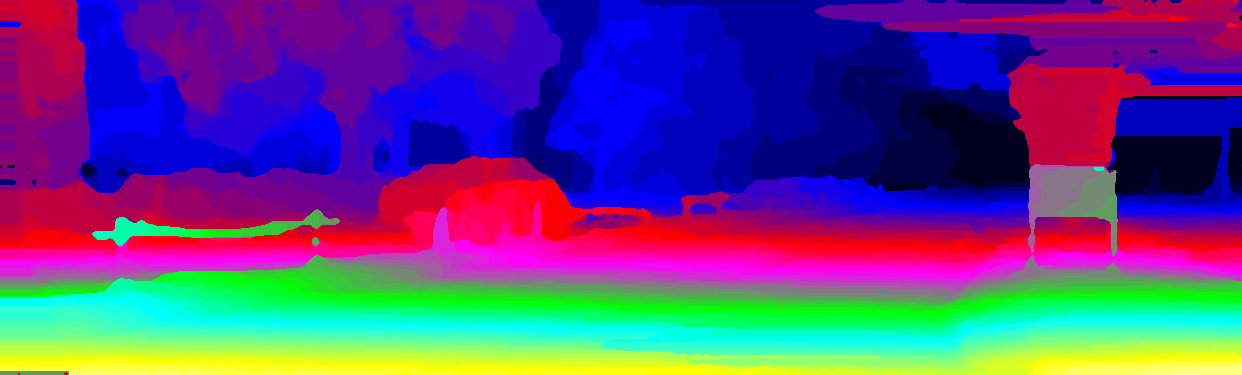}&
\includegraphics[width=5.0cm]{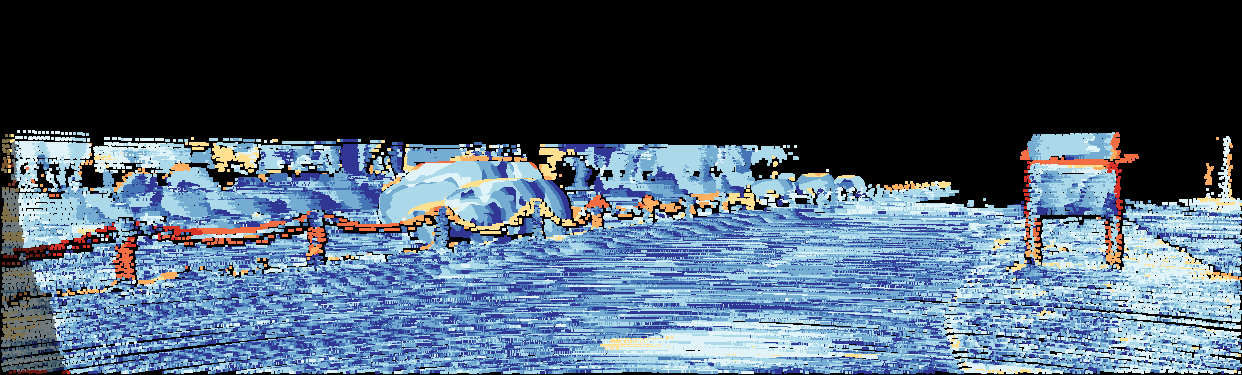}\\

\includegraphics[width=5.0cm]{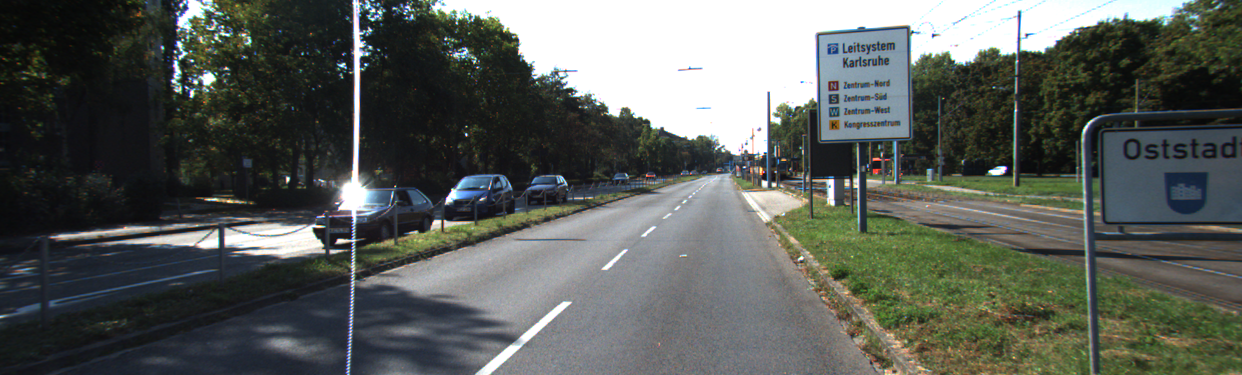}&
\includegraphics[width=5.0cm]{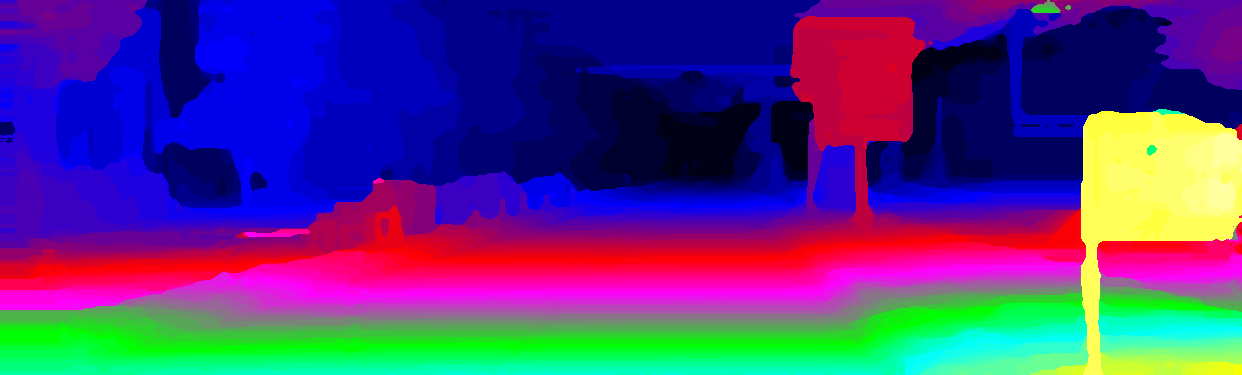}&
\includegraphics[width=5.0cm]{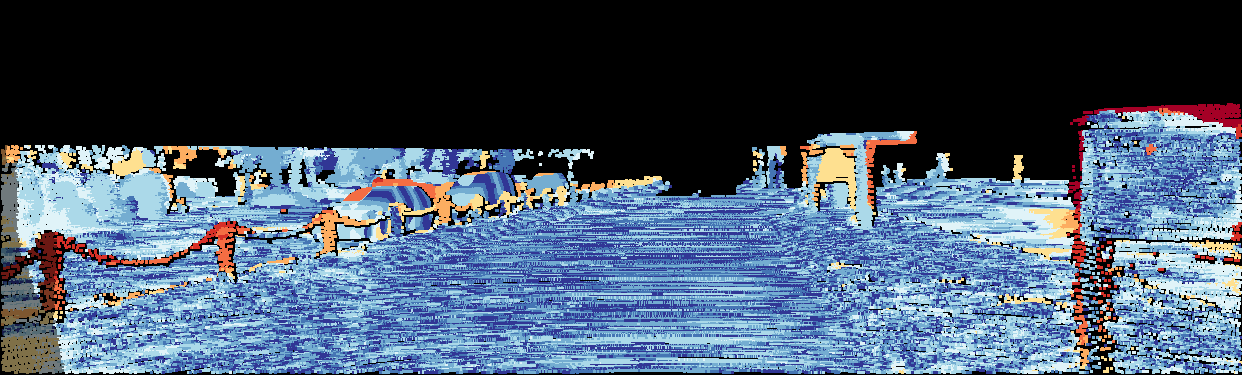}\\

\includegraphics[width=5.0cm]{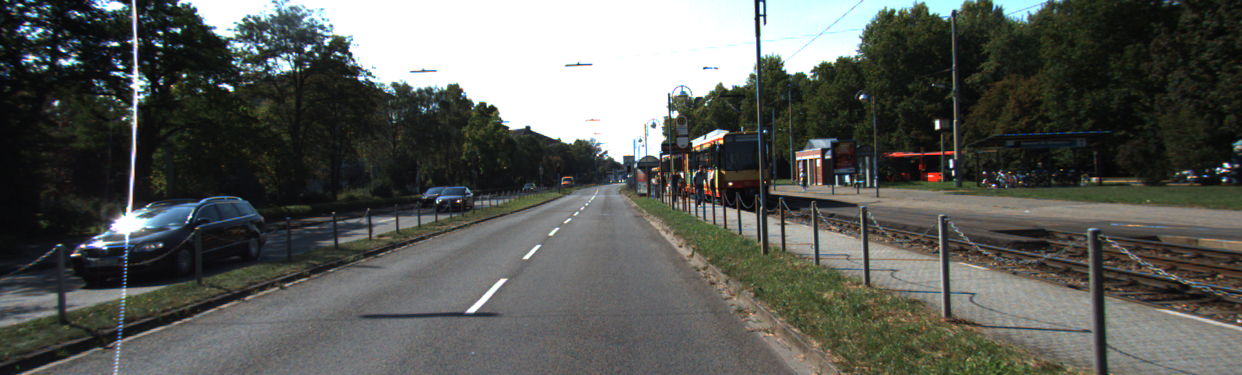}&
\includegraphics[width=5.0cm]{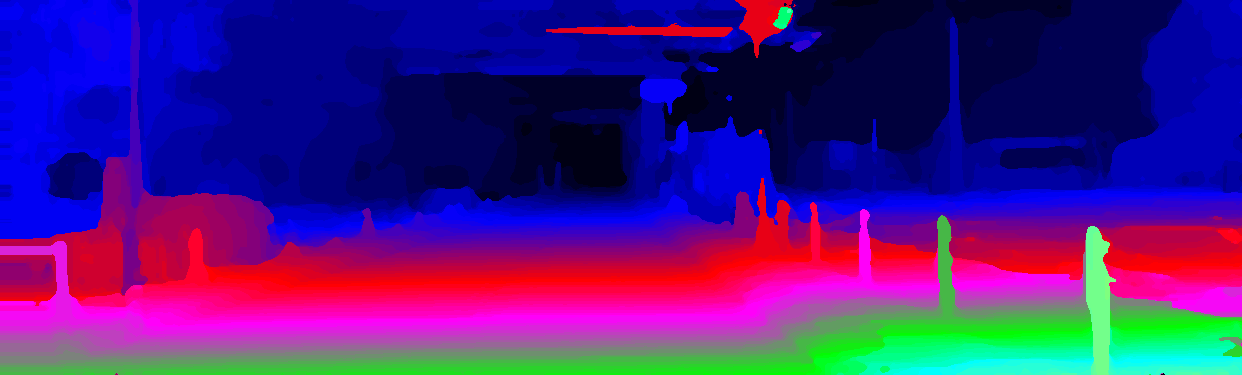}&
\includegraphics[width=5.0cm]{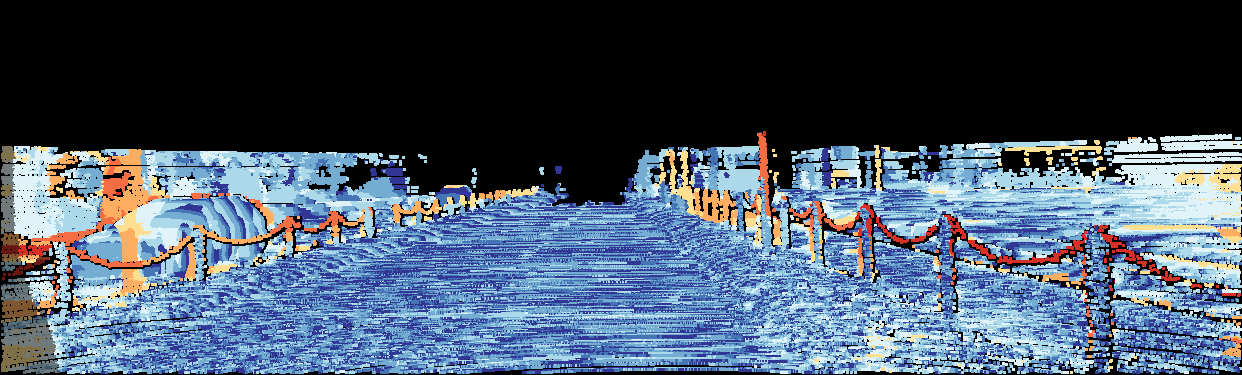}\\

\includegraphics[width=5.0cm]{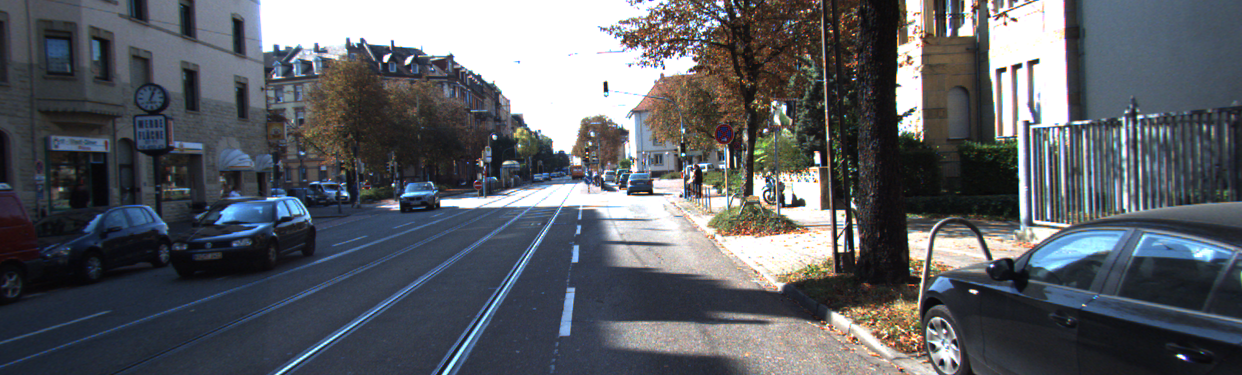}&
\includegraphics[width=5.0cm]{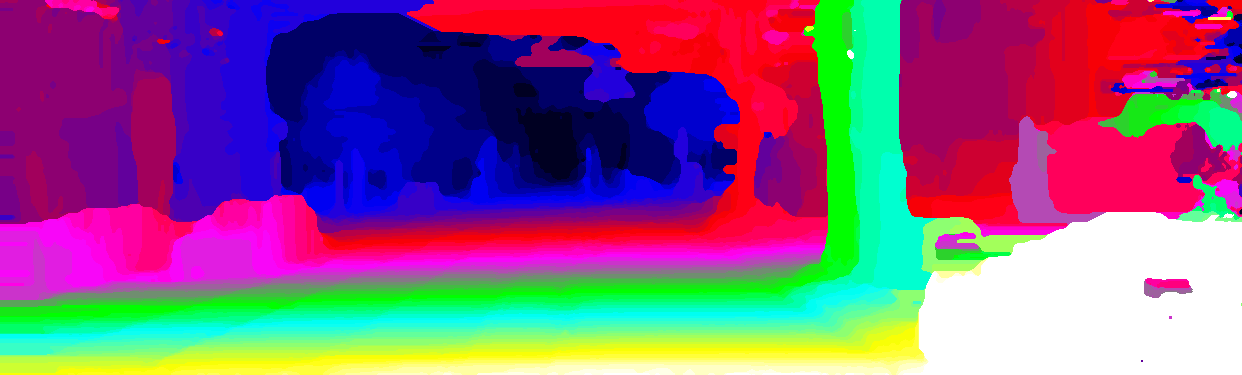}&
\includegraphics[width=5.0cm]{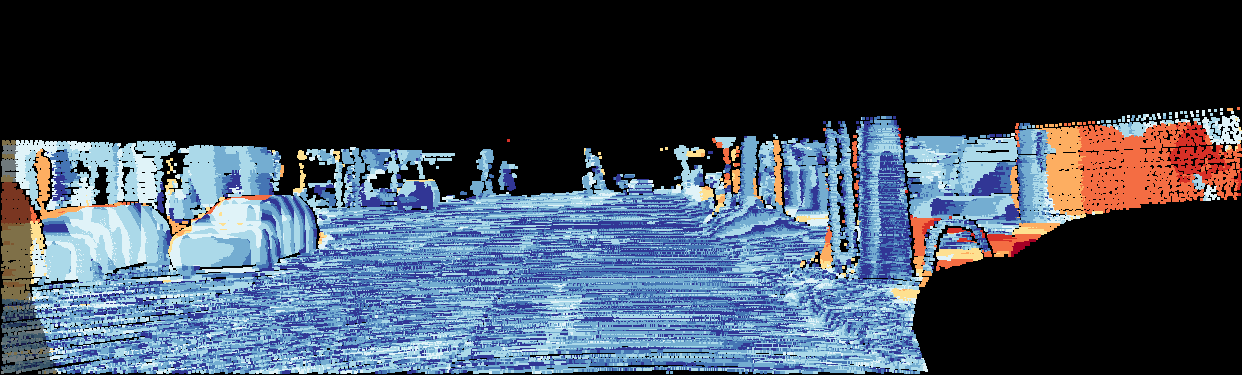}\\

\includegraphics[width=5.0cm]{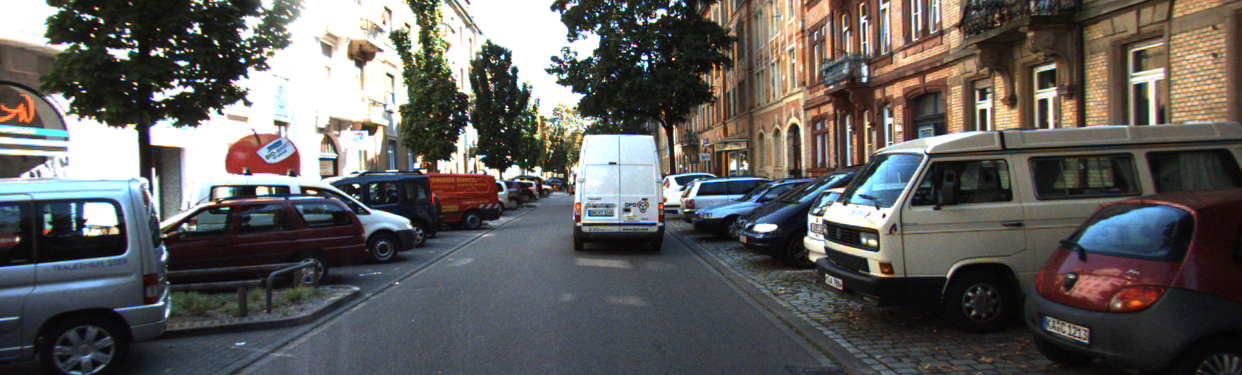}&
\includegraphics[width=5.0cm]{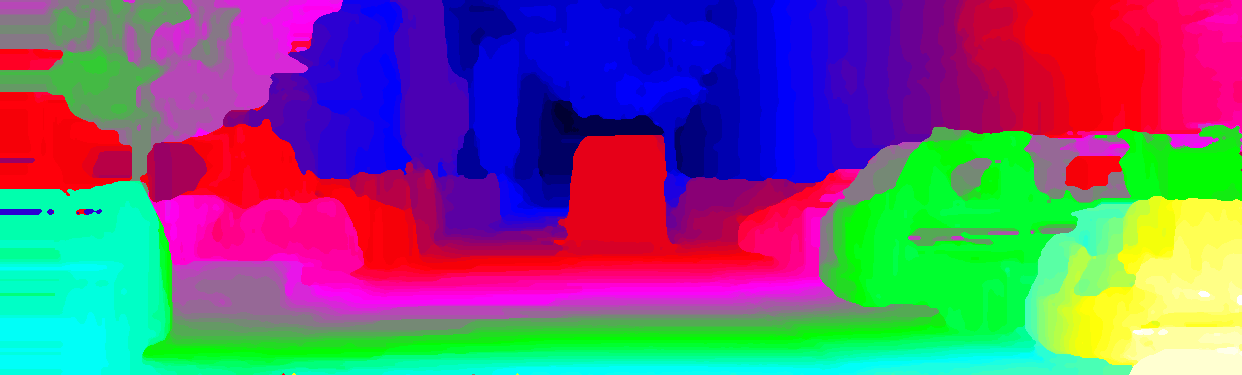}&
\includegraphics[width=5.0cm]{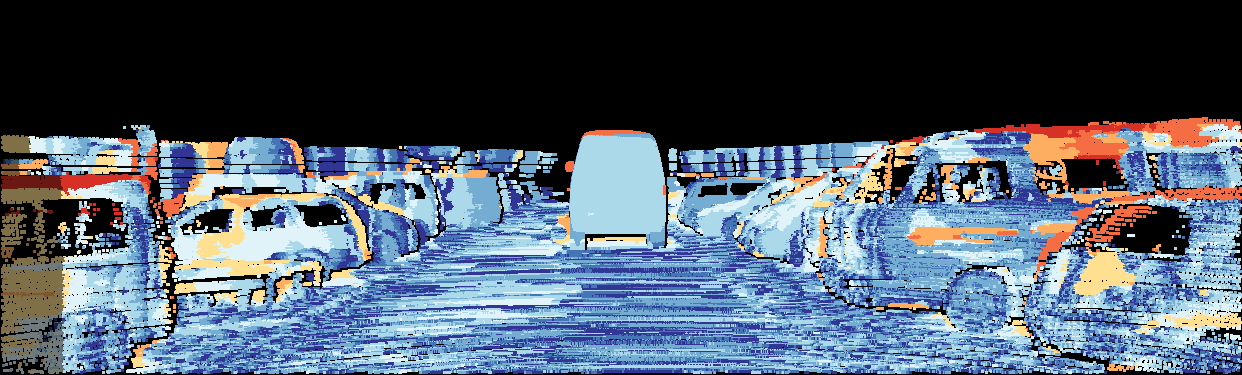}\\

\includegraphics[width=5.0cm]{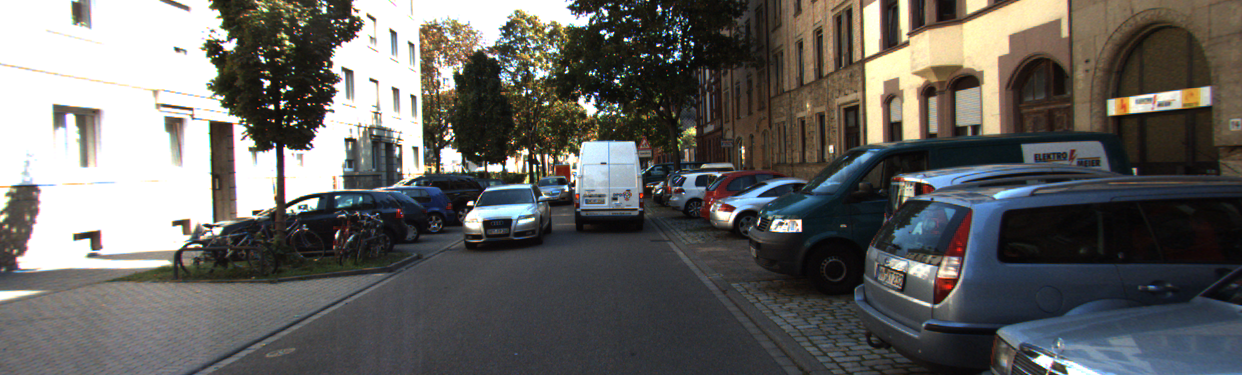}&
\includegraphics[width=5.0cm]{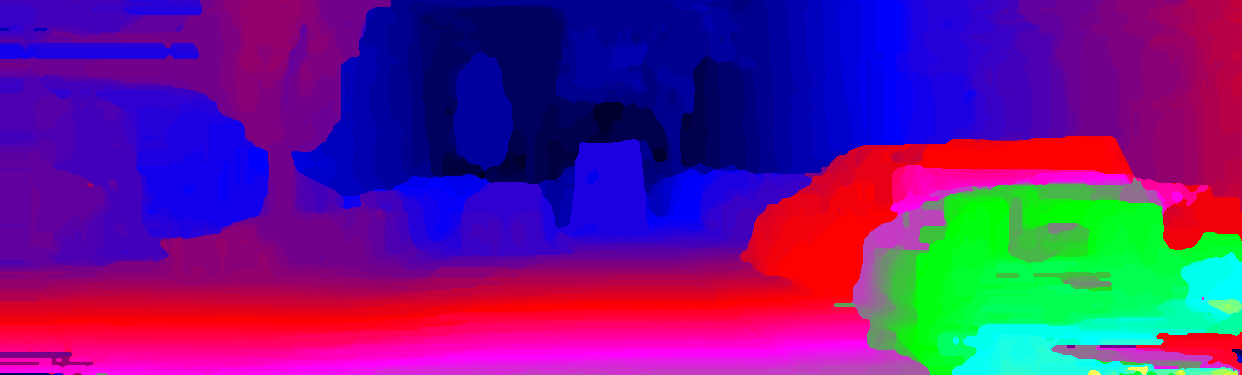}&
\includegraphics[width=5.0cm]{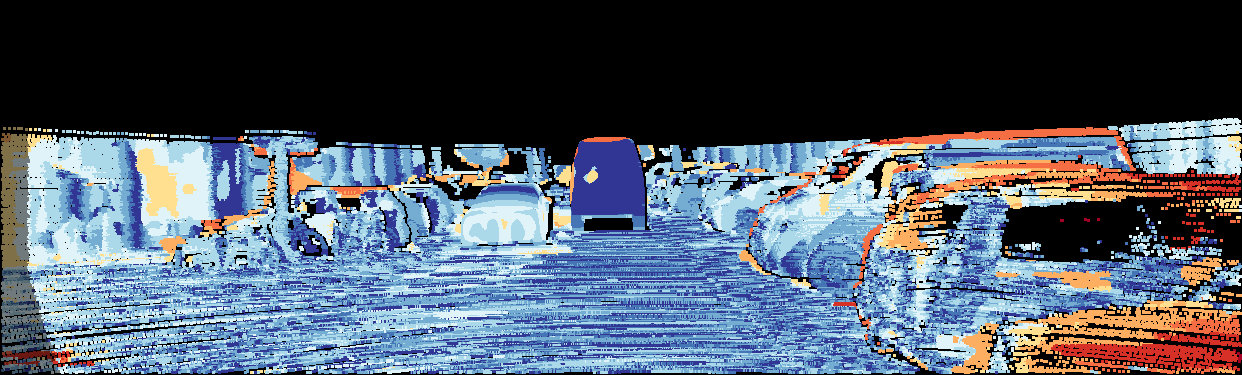}\\

\includegraphics[width=5.0cm]{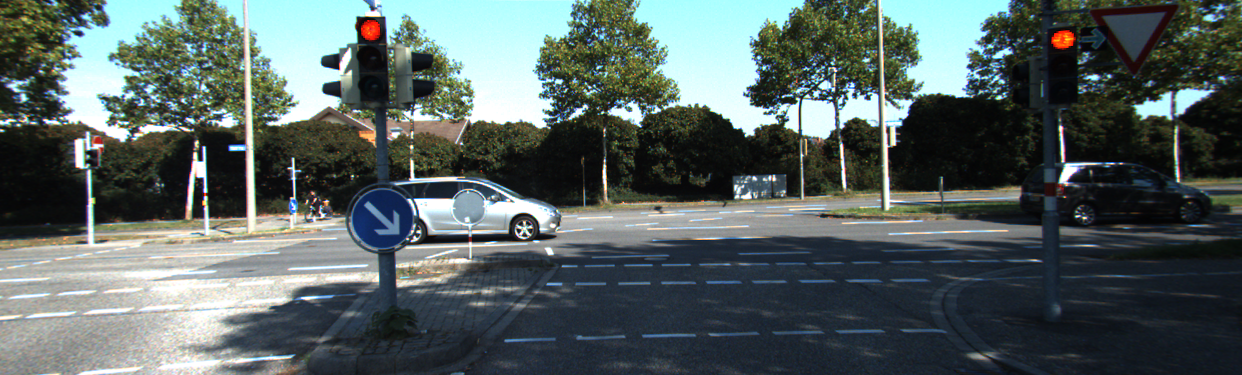}&
\includegraphics[width=5.0cm]{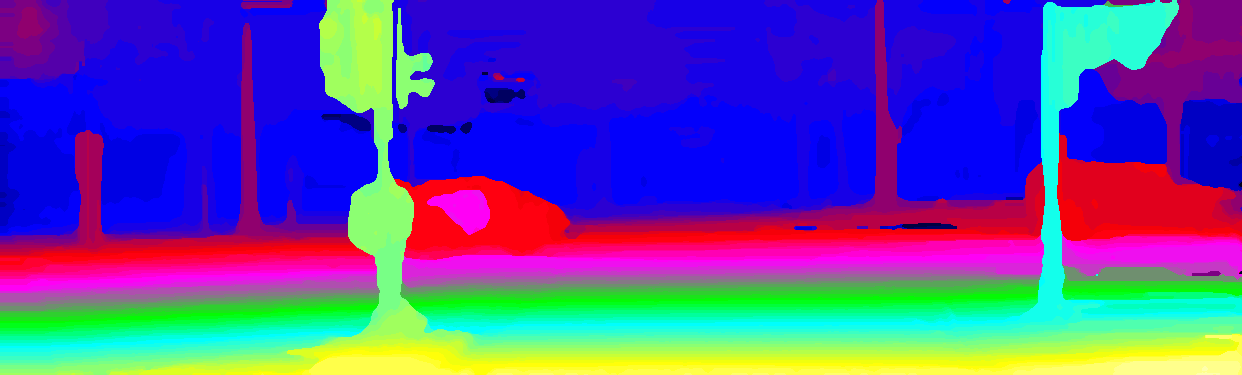}&
\includegraphics[width=5.0cm]{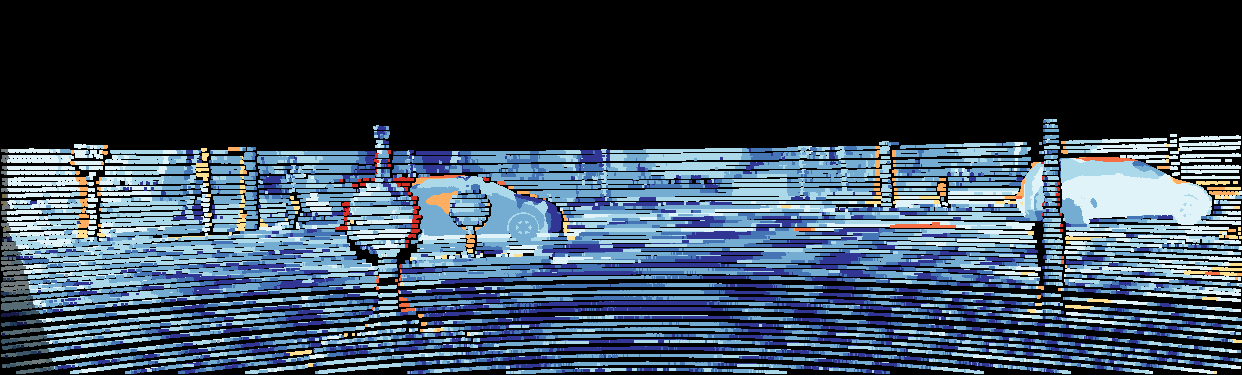}\\
\end{tabular}
\end{small}
\caption{KITTI 2015 test set: (left) original image, (center) disparity field, (right) stereo errors.}
\label{fig:kitti_results}
\end{figure*}

\subsection{Runtime}

The learning was implemented using the combination of Theano \cite{team2016theano} framework and fast CUDA kernels for the AD-census cost volume computation. The test time evaluation framework combines the edge detector implemented using CuDNN library with CUDA implementation of the the domain transform.
We measure the runtime of our implementation on a PC with NVIDIA GeForce GTX Titan X GPU. Training takes about 4-5 hours. 

\begin{table}
\begin{center}
  \begin{tabular}{c | c | c}
    \hline
    \textbf{stage}  & \textbf{\# of calls} & \textbf{total runtime, msec} \\ \hline
    data term & 2 & 3 \\ \hline
    CNN & 1 & 9 \\ \hline
    domain transform & 2 &  19 \\ \hline
    WTA & 2 &  1\\ \hline
    left-right check & 1 & 2\\ \hline
    \textbf{total} &  & \textbf{34}\\ \hline
  \end{tabular}
\caption{Runtime for our parallel implementation of the learned domain transform cost aggregation algorithm for KITTI 2015 dataset images on NVIDIA GeForce GTX Titan X GPU. The stereo algorithm is run twice since the left-right check is performed. The domain transform weights are predicted by a single forward pass of the CNN for a batch containing 2 images. Most of the time is consumed by domain transform and the edge detector.}
\label{tab:runtime}
\end{center}
\end{table}

The runtime of our method across the pipeline stages can be observed in the table \ref{tab:runtime}. The total run-time is 34 msec (29 frames per second) runtime per image pair including overhead associated with the left-right check. The greatest fractions of the run-time are spent for the domain transform and the edge detection CNN forward computation.   

\subsection{Quantitative results}

Our method achieves $6.34\%$ error on the KITTI 2015 data set. The predicted disparity fields for the first images of the test set can be observed on fig. \ref{fig:kitti_results}. The most notable challenges for the method arguably are dense ground truth labels for the car bodies. The method also fails to predict disparity for some thin curved regions that are not aligned with the vertical or horizontal axes. Overall it is able to produce correct labels for most of the labels which correspond to slanted surfaces, e.g. the road surfaces.

\section{Conclusion} \label{section_conclusion}

We proposed a new method of computing dense stereo correspondences using convolutional neural network trained to aggregate the cost volume. The methods is based on the multi-scale edge detector using to extract relevant edges and the domain transform trained to perform the cost volume aggregation. The method was evaluated using KITTI 2015 data set and achieves $6.34\%$ error at the 29 frames per second rate (see fig. \ref{fig:kitti_results}). 

Our approach can be extended to incorporate other cost aggregation approaches such as semi-global matching~\cite{hirschmuller2005accurate} that can also be unrolled to recurrent neural networks for training. We are currently investigating this approach, as it can potentially lead to more accurate aggregation, especially at slanted surfaces. 

\bibliographystyle{IEEEtran}
\bibliography{root}

\end{document}